\documentclass[10pt,conference,a4paper]{IEEEtran}

\usepackage[pdftex]{graphicx}
\usepackage{hyperref} 
\usepackage{color}
\usepackage{todonotes}
\usepackage{ulem}

\ifCLASSINFOpdf
\else
\fi

\definecolor{blue}{rgb}{0.00,0.00,1.00}

\begin{document}
%
\title{Learning to Rank for Active Learning: A Listwise Approach}

\author{\IEEEauthorblockN{Minghan Li\IEEEauthorrefmark{1}\IEEEauthorrefmark{2}, Xialei Liu\IEEEauthorrefmark{1}, Joost van de Weijer\IEEEauthorrefmark{1}, Bogdan Raducanu\IEEEauthorrefmark{1}}
\IEEEauthorblockA{\IEEEauthorrefmark{1}Computer Vision Center, Universitat Autonoma de Barcelona, Barcelona, Spain\\ \IEEEauthorrefmark{2}Universite Grenoble Alpes, Grenoble, France\\
Email: \{minghan, xialei, joost, bogdan\}@cvc.uab.es}

}


%


\maketitle

\begin{abstract}
Active learning emerged as an alternative to
alleviate the effort to label huge amount of data for
data-hungry applications (such as image/video indexing and retrieval, autonomous driving, etc.). The goal of active learning is
to automatically select a number of unlabeled samples for annotation (according to a budget), based on an acquisition function, which indicates how valuable a sample is for training the model.
The learning loss method is a task-agnostic approach which attaches a module to learn to predict the target loss of unlabeled data, and select data with the highest loss for labeling. In this work, we follow this strategy but we define the acquisition function as a learning to rank problem and rethink the structure of the loss prediction module, using a simple but effective listwise approach. Experimental results on four datasets demonstrate that our method outperforms recent state-of-the-art active learning approaches for both image classification and regression tasks.
\end{abstract}


%
\IEEEpeerreviewmaketitle

\section{Introduction}

There are many applications nowadays, such as image/video indexing and retrieval, autonomous driving, etc. which require a huge amount of labeled data. Manual annotation of this data is time consuming and prohibitively expensive since it involves human resources \cite{settles2012active}. As a result, active learning emerged as an alternative to make this process more manageable.

Active learning attempts to overcome the labeling bottleneck by automatically selecting the most valuable data to be annotated by human experts. Active learning assumes the existence of a small labeled dataset and a fixed budget to annotate the unlabeled samples. The goal of active learning is to automatically select a number of unlabeled samples (according to the budget) for annotation, based on an acquisition function (also known as ‘query function’) which indicates how valuable a sample is for training the model. The underlying assumption is that some data samples provide more valuable information than others, so that when labeled and used for training, they improve the model performance by decreasing the number of annotations. 
In this way, the active learner aims to achieve high accuracy using as few labeled instances as possible, thereby minimizing the cost of obtaining labeled data. Active learning has been successfully applied to several computer vision applications, such as: image classification
\cite{joshi2012classif,gavves2015classif}, object
detection \cite{zolfaghari2019temporal,aghdam2019active}, Visual Question Answering (VQA) \cite{lin2017vqa}, remote
sensing \cite{deng2018remotesensing} and action localization \cite{heilbron2018action}.

\begin{figure}[!t] \centering \includegraphics[width=3.2in]{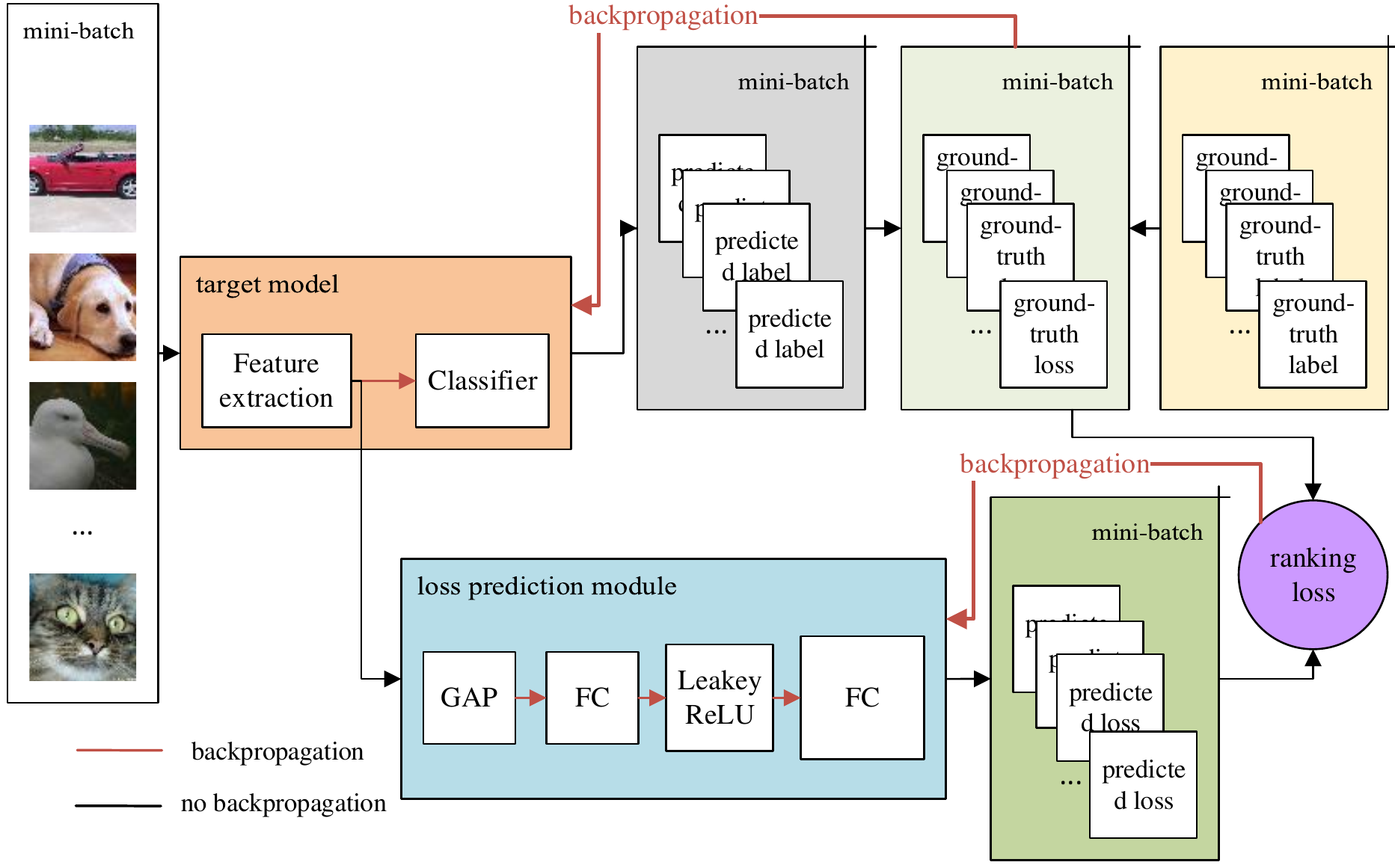}
\caption{The whole architecture of the proposed active learning algorithm.}
\label{archi} 
\end{figure}

How to define and select the most valuable data - i.e. the query strategy - is the main research topic of active learning. 
In the current work, we follow the idea of learning to predict a loss from \cite{yoo2019learning} as this method is task-agnostic and effective. This method learns to predict the loss of unlabeled input sample, and uses the predicted loss as a measure of uncertainty. Besides, unlike for instance the entropy method, which only suits classification problems, the learning loss method suits a variety of loss based deep learning problems, e.g. image  classification, object detection and human pose estimation.

The overall architecture of the proposed method is depicted in Fig.~\ref{archi}, for an image classification problem, but it could be easily adapted for regression problems as well. The motivation for training the loss prediction module (the blue box) is to minimize the errors of predicted losses (dark green box) and ground-truth losses (light green box). A ground-truth loss is calculated by a loss function using the predicted label (gray box) of target model (orange box) and the ground-truth label (yellow box).

However, the learning loss based active learning problem is actually a ranking problem (the purple circle). We clarify this aspect in subsection \ref{whylearnrank} and demonstrate that the loss prediction module should be trained by minimizing the ranking error. 

Although our approach has been inspired by \cite{yoo2019learning}, it is different in the following aspects (more details in subsection \ref{archisubsection}):
(i) from the target model (orange box) we extract only the features of the last convolutional layer to be processed by the rest of the pipeline, since it showed to improve the performance;
(ii) in the loss prediction module (blue box) we used an improved, more explicit method to rank the predicted losses which takes into account the list (of losses) structure
(iii) regarding training, we stop the  ranking loss  gradient to backpropagate to  the target  model  and  we separate  the  two  losses (ranking loss and the target loss), so  the loss  prediction  module  and  target model  are  trained  separately. This is better, because the gradient from the loss prediction module does not influence the target model.

We validate our proposed framework on four datasets: CIFAR-10 \cite{krizhevsky:2009} and CelebA \cite{liu2015deep} for classification tasks, and MPII dataset \cite{andriluka20142d} for human pose estimation and ShanghaiTech Part\_B dataset \cite{zhang2016single} for crowd counting for regression tasks. Experimental results demonstrate that our algorithm, learning to rank for active learning (L2R-AL), outperforms state-of-the-art methods such as: core-set \cite{sener2018active}, learning loss for active learning (LL4AL) \cite{yoo2019learning}, entropy method \cite{li2013adaptive} and Variational Adversarial Active Learning (VAAL).

To summarize, the main contributions are:
\begin{itemize}
\item we demonstrate the learning loss based active learning method is actually a learning to rank problem 
\item we use an improved ranking method for predicted losses (the listwise approach) 
\item we show that although for classification tasks entropy method dominates other active learning approaches (including ours), for regression tasks our approach holds the best performance when compared with current state-of-the art methods
\end{itemize}

The paper is organized as follows. The related work is presented in Section \ref{sec2}. Section \ref{sec3} details the proposed active learning algorithm. Section \ref{sec4} shows the experiment settings and results. Section \ref{sec5} concludes our paper.

\section{Related Work}
\label{sec2}
\subsection{Active Learning}

Most of the active learning strategies are pool-based approaches, which can be further divided into the following categories, depending on the query function: informativeness
\cite{guo2010active,cai2014active}, representativeness \cite{saito2015active,sener2018active}, hybrid \cite{huang2014active,yang2018activea} and performance-based \cite{gu2012active,fu2018active,yang2018activeb}. 

Among all the above approaches, informativeness-based approaches are the most successful ones, with uncertainty being the most used selection criteria used in both bayesian \cite{gal2017active} and non-bayesian frameworks \cite{yang2015active}. The entropy method \cite{li2013adaptive,joshi2009multi} calculates the entropy value of class posterior probabilities to define uncertainty, and data with the highest entropy is viewed as the most uncertain. Despite the query strategy is very simple, this method performs remarkably well for classification problems. However, this method is not suitable for regression problems and people need to design specific uncertainty metrics as shown in \cite{yoo2019learning}. 

The query-by-committee \cite{seung1992query,beluch2018power} is another popular active learning strategy, which alleviates many disadvantages of uncertainty sampling. For instance, uncertainty sampling tends to be biased towards the actual learner and it may miss important examples which are not in the sight of the estimator. The committee issues multiple hypotheses and the instance with highest consensus is viewed as the most informative. This motivation is simple and clear, however, for current DNNs, training a committee has high computational cost. 

In the era of big data and deep learning, it has been proven in \cite{sener2018active} that classical approaches as mentioned earlier, do not scale well to large datasets. Therefore, recently, the attention has been shifted
towards deep active learning, with adversarial strategies being one of the most popular techniques \cite{deng2018adversarial}.

The authors of core-set approach \cite{sener2018active} 
define the problem of active learning as core-set selection where the problem becomes a K-center problem which can be solved by using a greedy approximation method.
Empirical results showed state-of-the-art performance,
but the query strategy is computationally expensive.

Recently, \cite{yoo2019learning} has introduced a novel and task-agnostic active learning method.
The authors attach a loss prediction module to the target network, and learn to predict target model's losses of unlabeled samples. The data with highest predicted loss is viewed as the most uncertain and informative. The experimental results demonstrate that their method consistently outperforms the previous methods over the tasks. 
They learn the loss prediction module by considering the differences between pairs of loss predictions. However, the batch size of their method must be an even number and they only consider the neighbouring data pairs, neglecting the overall list relationship. 

The VAAL (Variational Adversarial Active Learning) approach \cite{sinha2019variational} 
learns a latent space using a variational autoencoder (VAE) \cite{kingma2013auto} and an adversarial network trained to discriminate between labeled and unlabeled data. 
Samples predicted as “unlabeled” with the lowest confidence is sent to the oracle. They demonstrate state-of-the-art results.
However, training the VAE and the discriminator requires high computational cost, and the hyperparameters may be sensitive.

In our paper, our active learning algorithm is based on the learning loss strategy of \cite{yoo2019learning}. Instead of trying to minimize mean square error of predicted losses and ground-truth losses, we define the problem as a learning to rank problem. We use a listwise approach to train the loss prediction module. Besides, we have further studied the architecture of loss prediction module. The detailed architecture of our algorithm can be seen in the next section.

\subsection{Learning to Rank}
\label{relateLearning}
The central problem of many tasks in information retrieval (IR) and natural language processing (NLP) is ranking, including document retrieval, question answering, meta-search, online advertisement, collaborative filtering, machine translation and so on \cite{li2011learning}.
Learning to rank means performing a ranking using machine learning techniques. \cite{lv2011learning} learns to identify a ranked list of related news articles which the user would like to read afterwards. 
Besides, learning to rank techniques are central to question answering systems for questions are matched
against an extensive database to find the most relevant answer \cite{tay2017learning}. Furthermore, \cite{xuan2014learning} uses learning to rank for fault localization in software debugging.

Learning to rank has two components: a learning system and a ranking system \cite{li2011learning}. In the learning system, for each request, there is a set of offerings  and there is a true ranking list on the offerings. The ranking system receives a subset of new offerings and assigns scores to them, where the system uses the ranking model trained by the learning system. Then the ranking list is obtained with the scores.

The authors of \cite{liu2009learning} group learning to rank problems into three approaches: the pointwise approach,
the pairwise approach, and the listwise approach. The pointwise approach assumes that each instance in the training data has a numerical or ordinary score, then it can be approximated by a regression problem: given a single query, predict its score. In the pairwise approach, ranking is transformed into a pairwise classification or pairwise regression. A major limitation of the pointwise and pairwise ranking approaches is that the group structure is ignored \cite{li2011learning}.  The listwise approach instead tries to optimize the value of an evaluation metric. The most commonly used metrics include: mean average precision (MAP), Spearman's rank correlation \cite{dodge2008concise}, normalized discounted cumulative gain (NDCG) \cite{chakrabarti2008structured}, etc. The listwise approach is difficult in the context of deep learning end-to-end architecures because most of the metrics are not differentiable with respect to ranking model's parameters, so surrogate functions are used. In practice, listwise approach often outperforms pairwise and pointwise approaches, and this is supported by a large scale of experiments \cite{tax2015cross}.

In this paper, we will demonstrate that the loss prediction based active learning algorithm actually is a learning to rank problem, which can be seen in subsection \ref{whylearnrank}. The proposed active learning algorithm adopts a listwise approach to train the loss prediction module. In the subsection \ref{sortdeep}, we will describe the surrogate method we use in the listwise approach.

\section{Learning to rank for active learning}
\label{sec3}

Let $f_{\Theta_{tg}}$ represents the target model, e.g., an image classification model or a regression model.
For a mini-batch of training data of size $d$, these training data has $d$ ground-truth labels $\{y_1, y_2, \dots, y_d\}$. The model
$f_{\Theta_{tg}}$ generates $d$ predicted labels $\{\hat{y}_1, \hat{y}_2, \dots, \hat{y}_d\}$, and we can calculate $d$ ground-truth losses $l_{\Theta_{tg}}=\{l_1, l_2, \dots, l_d\}$ with these data. For learning loss based active learning, let $f_{\Theta_{pre}}$ represent the loss prediction module which generates $d$ predicted losses $\hat{l}_{\Theta_{pre}}=\{\hat{l}_1, \hat{l}_2, \dots, \hat{l}_d\}$.
During training, we train $f_{\Theta_{pre}}$ with the ground-truth losses, and during the querying of data for labeling by the oracle, a set of unlabeled instances with highest predicted losses are selected. The aim is to learn the parameters of the loss prediction module, $\Theta_{pre}$, in order to predict higher loss scores for unlabeled instances which it is more uncertain about.

\subsection{Why Learning to Rank}
\label{whylearnrank}
The query strategy of \cite{yoo2019learning} consists of choosing a set of samples with high predicted losses. Learning the loss prediction module by mean square error (MSE) with the ground-truth losses is a simple idea. However, the authors said they failed to learn a good loss prediction module with MSE: the scale of the real loss decreases overall with target model's learning, so the loss prediction module would adapt roughly to the scale rather than fitting to the exact value. Their solution is to calculate the loss of loss prediction module by comparing pairs of values, i.e. they adopt a pairwise ranking approach. For a mini-batch whose size is $d$, they make $d/2$ data pairs and consider the difference between each pair of predicted losses and ground-truth losses thus discarding the overall scale changes \cite{yoo2019learning}.

\begin{figure}[!t] \centering \includegraphics[width=3.2in]{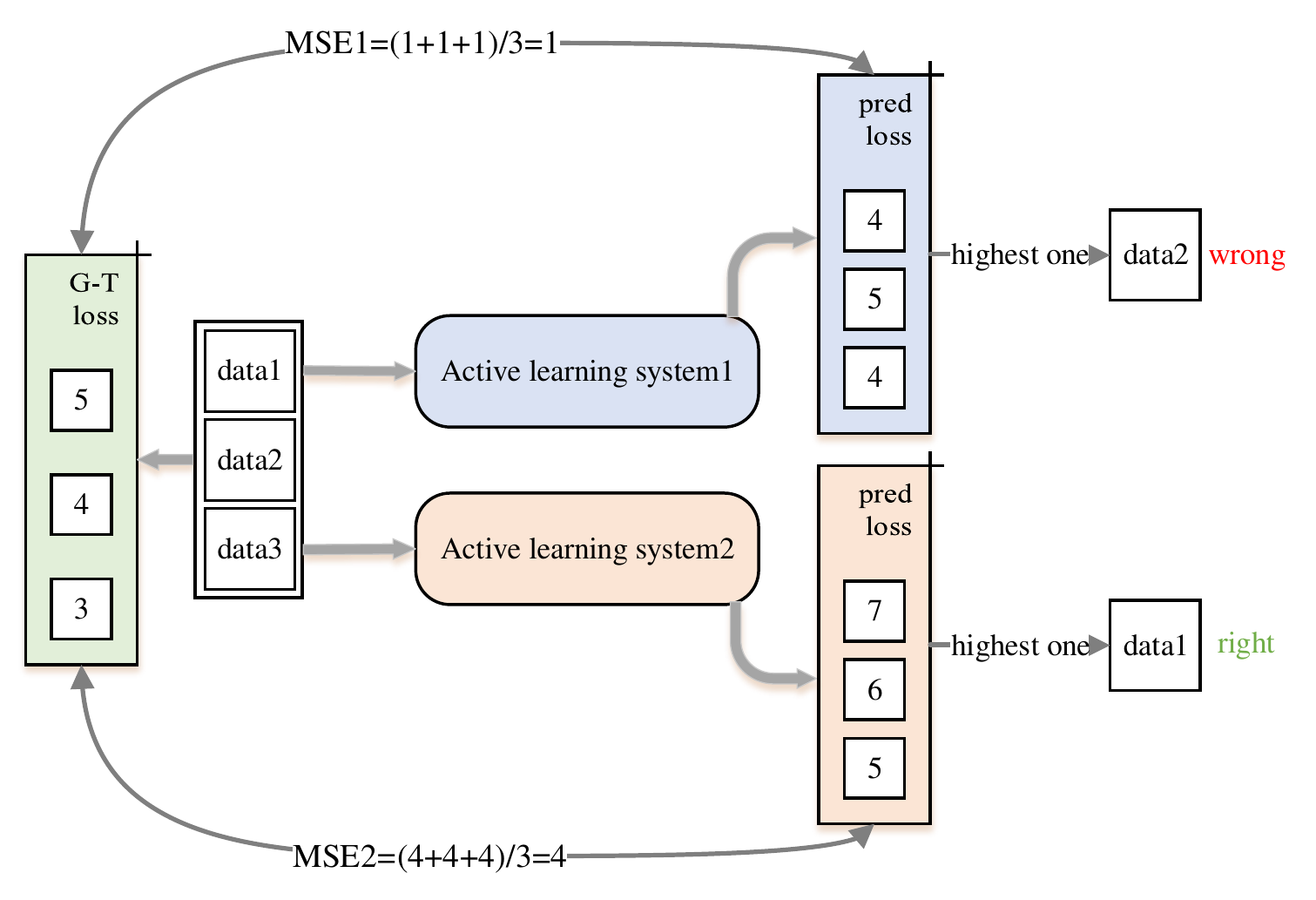}
\caption{An example of why the module should be trained with rank.}
\label{whyrank} 
\end{figure}

However, we demonstrate that the problem is not only that the scale of loss changes, but also that the loss prediction module should be trained with ranking loss. Fig.~\ref{whyrank} shows an example of why MSE is not optimal for training the learning loss module (G-T represents ground-truth). The `Active learning system 1' has lower MSE value but it recommends wrong data with highest uncertainty, whereas the `Active Learning system 2' recommends right data although its MSE value is higher. This is because `System 2' predicts the right ranks for the unlabeled data. This scenario motivates us to
use a better ranking scheme, based on the listwise approach.

\subsection{A Sorting Deep Net to Learn Ranking Loss Surrogates}
\label{sortdeep}
Many tasks are evaluated using non-differentiable metrics such as mean average precision or Spearman's rank correlation \cite{dodge2008concise}.
However, it is hard for us to use them as objective functions to train learning models for they are not differentiable. There are surrogate approaches but it is not easy to propose good surrogate functions \cite{engilberge2019sodeep}.

Recently, \cite{engilberge2019sodeep} proposes a method to learn to optimize such non-differentiable metrics. They use a deep neural network as a sorter to approximate the ranking function. It is pretrained with synthetic values and their ground-truth ranks. This sorter can then be combined with an existing model (e.g. loss prediction module) and converts the value list given by the model into ranking list. Then the ranking loss between the predicted ranks and ground-truth ranks can be calculated and backpropagated through the differentiable sorter and used to update the weights of the model.

\subsection{The Architecture of Proposed Algorithm}
\label{archisubsection}
As demonstrated above, we want to minimize the ranking error between the predicted losses and ground-truth losses. Let us consider $rk$ the ranking function which converts the ground-truth losses into ground-truth ranks $ rk(l_{\Theta_{tg}})=\{rk(l_1), rk(l_2), \dots, rk(l_d)\}$. The function $rk$ also converts the 
predicted losses into predicted ranks $rk(\hat{l}_{\Theta_{pre}})=\{rk(\hat{l}_1), rk(\hat{l}_2), \dots, rk(\hat{l}_d)\}$. 
For a mini-batch of input data with size $d$, the aim is to learn the parameters $\Theta_{pre}$ for:

\begin{equation} 
\label{error1} 
\min_{\Theta_{pre}}{Error(rk(l_{\Theta_{tg}}), rk(\hat{l}_{\Theta_{pre}}))}
\end{equation}
where $Error$ is a metric that calculates the error between two ranking loss lists.

\begin{figure}[!t] \centering \includegraphics[width=3.2in]{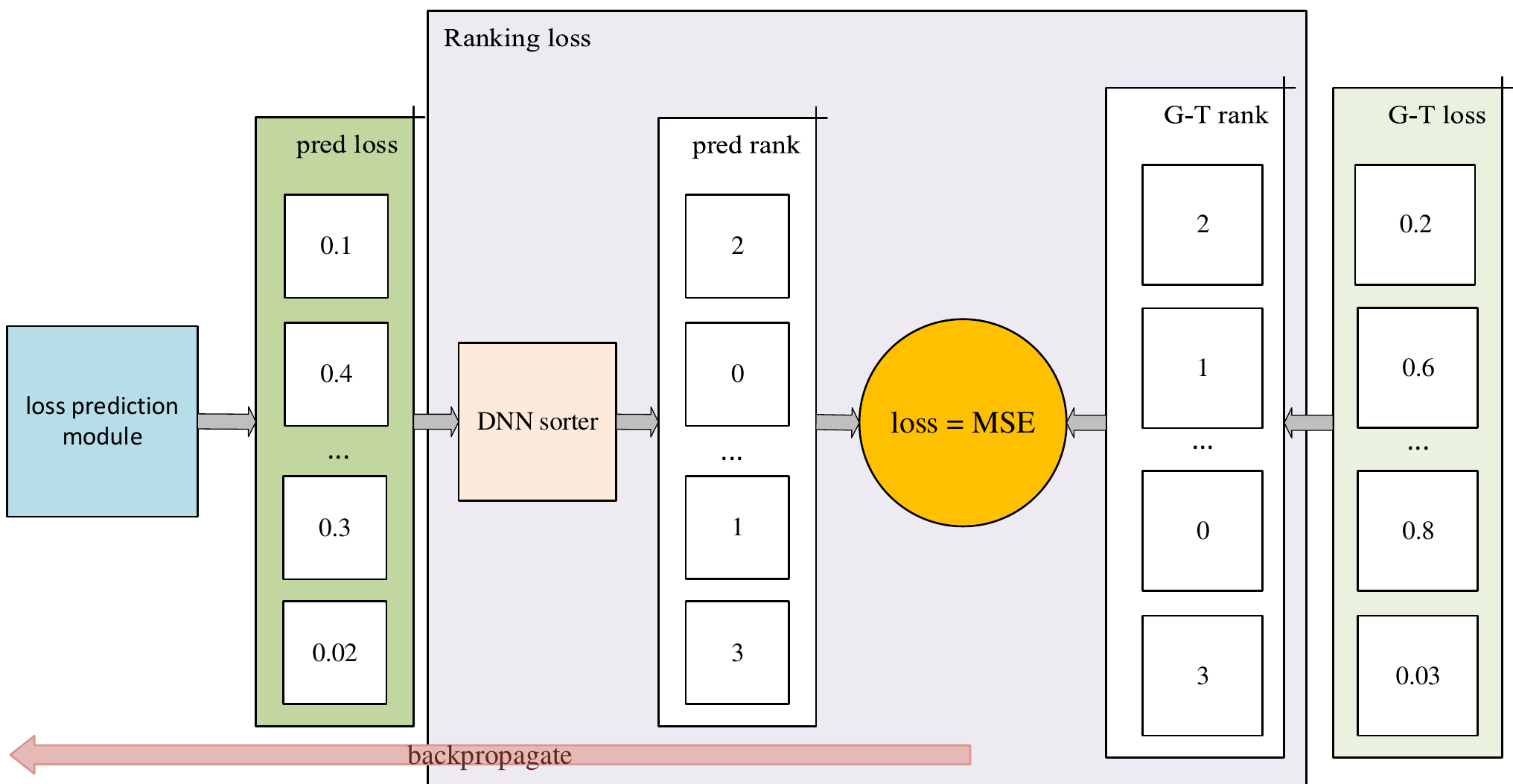}
\caption{A pretrained sorter is used to convert the predicted losses into predicted ranks. Thus the ranking loss is differentiable, which is equivalent to optimize Spearman's Rank correlation.}
\label{losspic} 
\end{figure}

However, the ranking function is not differentiable so the error as shown in (\ref{error1}) cannot be backpropagated to update $\Theta_{pre}$. As described in \ref{sortdeep}, \cite{engilberge2019sodeep} proposes a differentiable sorter to learn approximations of the rank vector $rk(\hat{l}_{\Theta_{pre}})$. In this paper, we adopt this method to active learning, aiming to minimize the error between predicted losses and ground-truth losses in the ranking space. Fig.~\ref{losspic} shows the solution to this problem.

Fig.~\ref{archi} shows the whole architecture of the proposed algorithm for image classification problems, where the predicted loss, ranking loss and ground-truth loss consist of Fig.~\ref{losspic}. The architectures for other kinds of problems are similar, i.e. for regression.

In \cite{yoo2019learning}, they take the features that are extracted between the mid-level blocks of the target model. However, we use the outputs of last block before the fully-connected layer from the target model as the features of input instances, as we find this method is more effective than using features from all mid-level blocks, and extracting and concatenating features from all blocks is not efficient for very deep neural networks. Besides, using the output of last block as feature is widely adopted in applications, e.g., image captioning \cite{xu2015show}.

The ranking loss backpropagates to the loss prediction module and updates its parameters $\Theta_{pre}$. 
In \cite{yoo2019learning}, the ranking loss and the target model's loss are combined with a hyperparameter, and the gradient from ranking loss backpropagates to the target model for some epochs. This will influence the target model's performance because the loss prediction module and target model are two different optimization problems. Besides, searching for an optimal hyperparameter is also difficult.

In this paper, we stop the ranking loss gradient to the target model and we separate the two losses, removing the hyperparameter, so the loss prediction module and target model are trained separately. The loss prediction module is only used to choose the most uncertain unlabeled instances to be labeled by the oracle. This is another difference with respect to \cite{yoo2019learning}.

The loss prediction module consists (see Fig.~\ref{archi}, the blue box) of a global average pooling (GAP) layer, a fully-connected layer, a Leaky ReLU activation function and another fully-connected layer. The GAP is used to decrease the feature scale, as many CNNs do, e.g., ResNet \cite{he2016deep} also contains a GAP before the fully-connected layer. We use a Leaky ReLU activation function as this is effective and converges faster. The two fully-connected layers are the core of this module, and they are learned to predict the losses of unlabeled instances by minimizing the ranking loss. \cite{yoo2019learning} also uses a GAP layer and a fully-connected layer to decrease the feature scale, then concatenate them to another fully-connected layer to predict the loss. 

\subsection{Loss Functions}
For a mini-batch with size $d$, the loss of target model is:
\begin{equation} 
\label{targetlossbatch} 
loss_{\Theta_{tg}}^{bat}=\frac{1}{d}\sum_{i=1}^d{l_i}
\end{equation}

For the loss prediction module, we use a pretrained differentiable sorter to convert the predicted losses to ranking list, and use MSE as the metric to calculate the error of the predicted ranks and ground-truth ranks. 
As shown in \cite{engilberge2019sodeep}, this method approximates the Spearman's rank correlation, which is a metric that measures the differences between two lists in ranking space.

For the predicted loss list $\hat{l}_{\Theta_{pre}}$ and the ground-truth loss list $l_{\Theta_{tg}}$, each with size $d$. The Spearman's rank correlation \cite{dodge2008concise} is defined as:
\begin{equation} 
\label{spearmanrank} 
r_s=1-\frac{6\|rk(l_{\Theta_{tg}})- rk(\hat{l}_{\Theta_{pre}})\|_2^2}{d(d^2-1)}
\end{equation}

The range of this metric is from -1 to 1. If the predicted ranks are same as the ground-truth ranks in every dimension, the value is 1, otherwise -1. Our aim is to maximize (\ref{spearmanrank}), this equals to:
\begin{equation} 
\label{minimi} 
\min_{\Theta_{pre}}{\|rk(l_{\Theta_{tg}})- rk(\hat{l}_{\Theta_{pre}})\|_2^2}
\end{equation}

As discussed, we use an pretrained differentiable sorter to approximate $rk(\hat{l}_{\Theta_{pre}})$. 
Let $f_{\Theta_{sort}}$ represent the sorter, then the ranking loss of $f_{\Theta_{pre}}$ becomes:
\begin{equation} 
\label{rankingloss} 
loss_{\Theta_{pre}}={\|rk(l_{\Theta_{tg}})- f_{\Theta_{sort}}(\hat{l}_{\Theta_{pre}})\|_2^2}
\end{equation}
Note that the sorter is pretrained independently on specific synthetic data. 

For a mini-batch with size $d$, then the ranking loss of this batch is:
\begin{equation} 
\label{rankinglossbatch} 
loss_{\Theta_{pre}}^{bat}=\frac{1}{d}\sum_{i=1}^d{(rk(l_i)-f_{\Theta_{sort}}(\hat{l}_i))^2}
\end{equation}

So maximizing the Spearman's rank correlation amounts to minimizing the loss in (\ref{rankinglossbatch}), and this is the mean square error (MSE). Here we can see as Fig.~\ref{losspic} shows, that the Spearman's rank correlation becomes differentiable and the loss prediction module $f_{\Theta_{pre}}$ can be trained by minimizing an MSE.

\subsection{Query Strategy}
In this work we adopt the standard methodology used in active learning literature, i.e. we divide the available labeling budget in cycles.
After the loss prediction module is trained in each cycle, it is used to select the unlabeled samples which are sent to the oracle for labeling. Since we are only interested in predicted losses, the query strategy does not require the ranking procedure.  Thus, for a given mini-batch, the loss prediction module outputs a list of predicted losses. The higher the loss is, the higher the rank is in this list. The query strategy of the proposed algorithm is choosing the instances with highest predicted losses, which equals to highest predicted ranks (i.e. highest uncertainty)

\section{Experiments}
\label{sec4}
In this section, we show the results of the proposed active learning algorithm on two image classification tasks using CIFAR-10 \cite{krizhevsky:2009}, and CelebA \cite{liu2015deep} datasets, and two regression tasks using MPII \cite{andriluka20142d} and ShanghaiTech Part\_B \cite{zhang2016single} datasets.

For all experiments, we choose the bidirectional GRU sorter to rank the losses\footnote{\url{https://github.com/technicolor-research/sodeep}}. We  train the GRU sorter with synthetic data according to the code of \cite{engilberge2019sodeep}. The synthetic training data consists of vectors of randomly generated scalars, associated with their ground-truth rank vectors. We train the sorter for 400 epochs, using a sequence length of 64, for the case of image classification task. For regression task, we use a sequence length of 32 for the MPII dataset and a sequence length of 4 for ShanghaiTech Part\_B dataset. Afterwards, we fix its parameters and use it to convert the predicted losses into a ranking list. 

We implement our algorithm and comparing methods in PyTorch \cite{paszke2017automatic}.

\subsection{Experiment on CIFAR-10}

\noindent
\textbf{Implementation details.} We use the 18-layer residual network (ResNet-18) \cite{he2016deep} as the target model for image classification. The batch size is 64. CIFAR-10 dataset contains 50000 training images and 10000 test images. Active learning algorithms begin from an initial labeled set and continue to be trained for several cycles. At each cycle the query strategy selects some unlabeled instances to be labeled by the oracle and add them to the labeled set. Afterwards, the target model is trained again.

In each active learning cycle, the active learning algorithms query instances for labeling from a random subset rather than the whole unlabeled pool. Subset size is set to 10000 for CIFAR-10. We begin the active learning cycle with 1000 labeled instances, and in each cycle, we continue to train the target model by adding 1000 labeled instances. We train the target model for 10 cycles.

We use the SGD optimizer to train the target model ResNet-18 of all the active learning algorithms for 150 epochs on training set, with initial learning rate of 0.1. After 120 epochs, the learning rate is decreased to 0.01. The momentum and the weight decay are 0.9 and 0.0005 respectively. We train the loss prediction modules of our algorithm and learning loss approach \cite{yoo2019learning} with Adam optimizer with a learning rate of 1e-3 for 150 epochs. The parameter $\alpha$ of Leaky ReLU is 0.01.

\noindent
\textbf{Comparing algorithms.} We compare the proposed ranking based active learning algorithm (L2R-AL) with three state-of-the-art approaches: core-set approach \cite{sener2018active}, VAAL \cite{sinha2019variational}, and Learning loss for Active Learning (LL4AL) \cite{yoo2019learning}, and two baselines: random method and entropy method \cite{li2013adaptive}. 

For core-set approach, we use the $K$-Center-Greedy algorithm in \cite{sener2018active} in accordance with \cite{yoo2019learning}. For VAAL approach, we use their official released code, and the VAE latent dimension is 32. We train the VAE with Adam optimizer with a learning rate of 5e-4 as the authors do.

We use the same augmentation scheme as shown in \cite{yoo2019learning}, including 32$\times$32 size random crop from 36$\times$36 zero-padded images, random horizontal flip, and normalize images using the channel mean and standard deviation vectors estimated over the training set.

\begin{figure}[!htbp] \centering \includegraphics[width=3.2in]{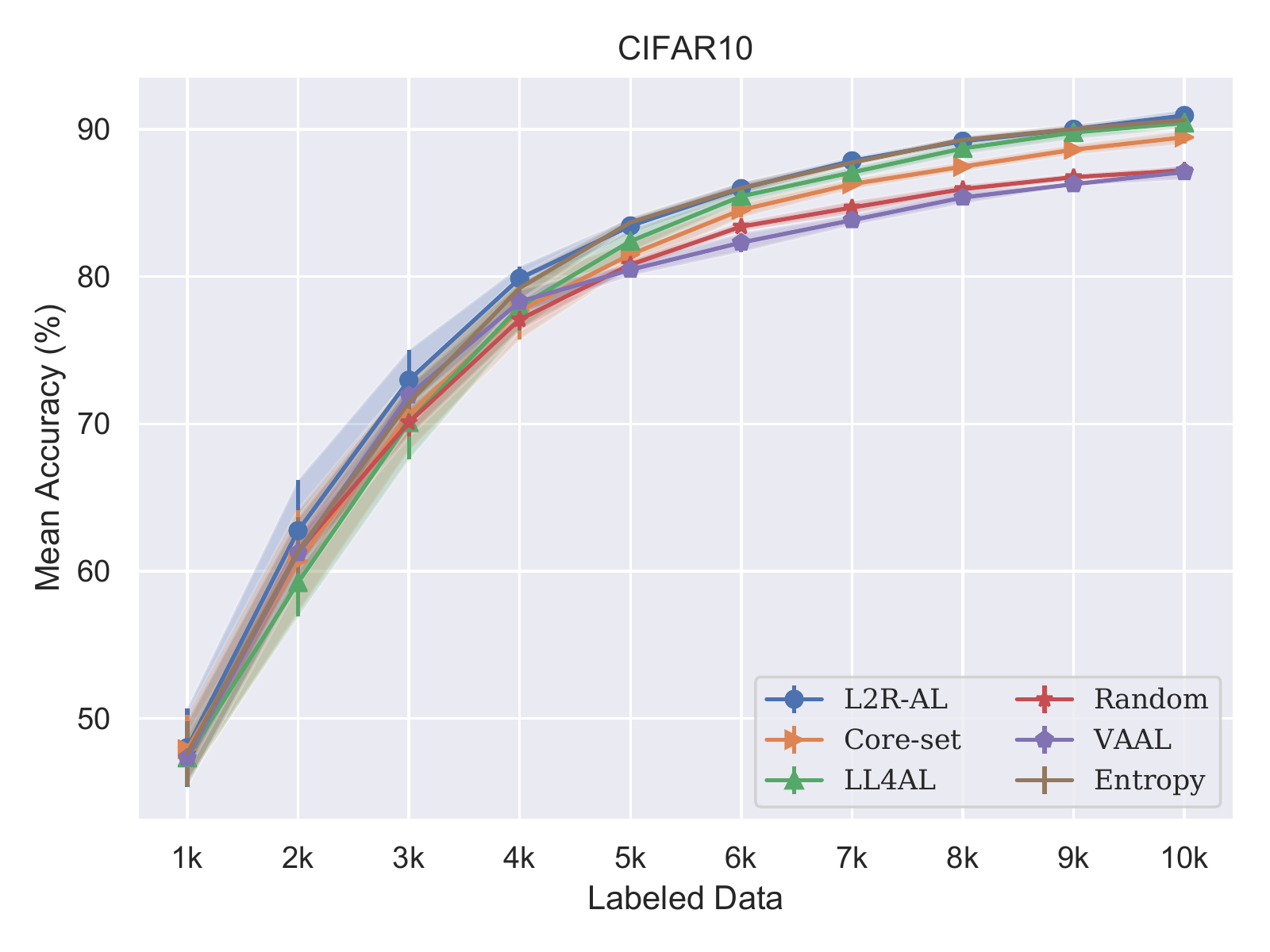}
\caption{Active learning results of CIFAR-10 image classification}
\label{cifar10} 
\end{figure}

\noindent
\textbf{Results.} Fig.~\ref{cifar10} shows the different active learning algorithms' average accuracies of 5 runs on the test set. For each run, different algorithms have the same random seed for selecting the initial labeled set and the subset in each cycle.

Our algorithm is noted as L2R-AL and it shows higher image classification accuracies than others. In the last active learning cycle, the random query strategy obtains a 87.72\% accuracy while the VAAL method achieves 87.11\%. We use the default hyperparameters of the VAAL official code but fail to get a better result.
LL4AL achieves a higher result than the core-set approach: 90.45\% and 89.47\% respectively. In the last several cycles, the entropy method performs similarly good as our algorithm. In the last cycle, the entropy method achieves 90.64\% and our algorithm achieves a 90.95\% result. The results shows that the proposed learning to rank active learning algorithm achieves better results than the other state-of-the-art comparing algorithms for CIFAR-10 problem.

\begin{figure}[!htbp] \centering \includegraphics[width=3.2in]{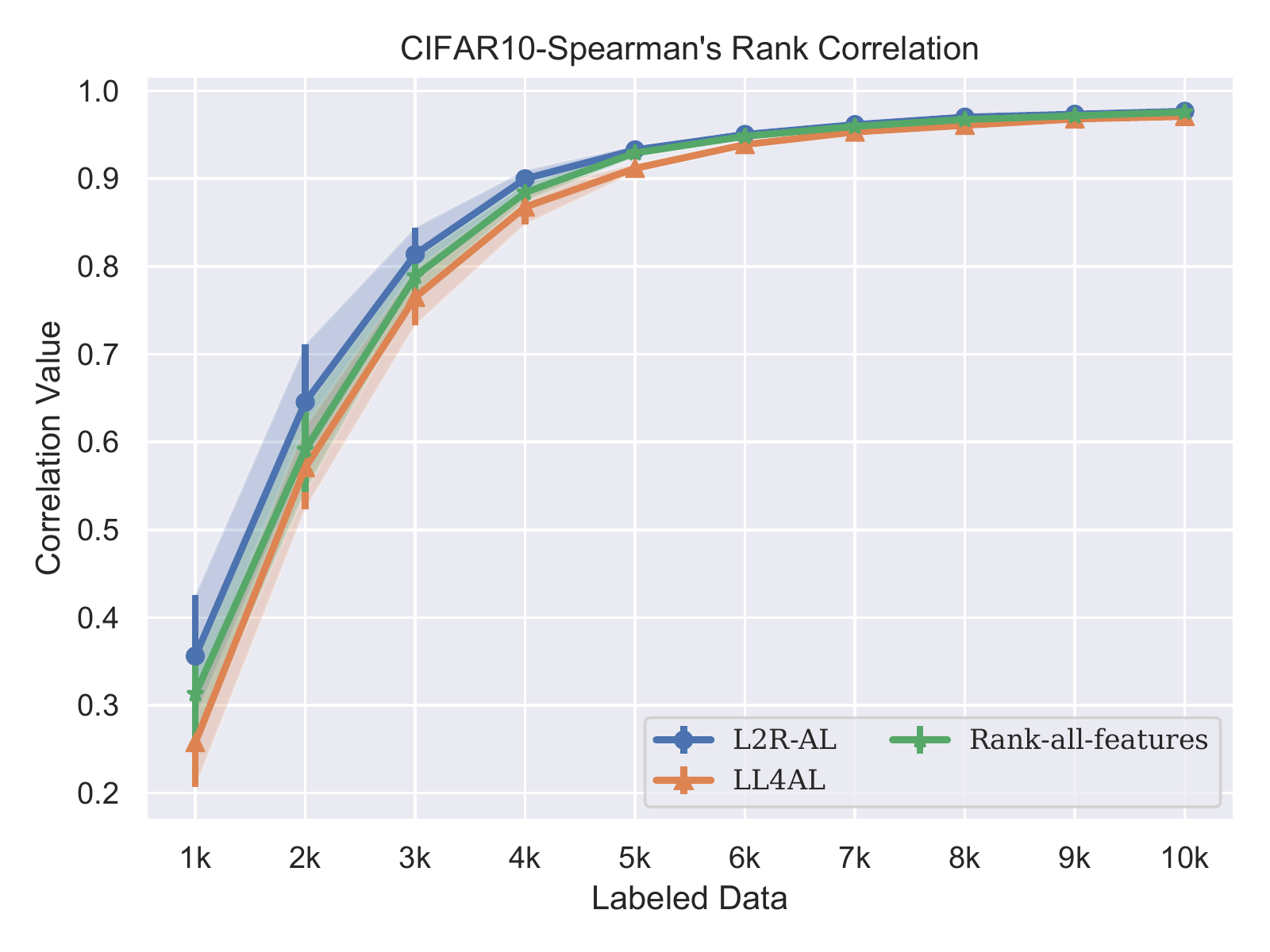}
\caption{The comparison of Spearman's rank correlation on CIFAR-10}
\label{cifar10Corr} 
\end{figure}

Besides, we want to show the ranking comparisons of our approach and LL4AL for each cycle. We use the Spearman's rank correlation \cite{dodge2008concise} to calculate the ranking effect of predicted losses and the ground-truth losses on the test set, as Fig.~\ref{cifar10Corr} shows. The `Rank-all-features' represents the active learning method that uses the same loss prediction module as the LL4AL approach with features from all mid-level blocks but trained with our proposed ranking method. We can see that our approach achieves the highest ranking result on the test set, especially in the beginning cycles. In the last cycle, it achieves 0.977 compared with 0.971 of the learning loss approach and 0.975 of the `Rank-all-features'. This result shows that the proposed L2R-AL not only has better loss function for training, but also has better architecture of loss prediction module.

\subsection{Experiment on CelebA}
\noindent
\textbf{Implementation details.} CelebA is a more challenging dataset which consists of 40 face attributes. This dataset contains 162770 training images and 19962 test images. The subset size is 20000. We choose the gender classification problem as image classification task using the male/female attribute. We resize all the images to 64$\times$64 pixels and normalize them.

We train the target model and loss prediction modules with SGD optimizer for 100 epochs with initial learning rate of 0.1. After 80 epochs, the learning rate is decreased to 0.01. 

\noindent
\textbf{Comparing algorithms.} We use the same comparing algorithms as the experiment on CIFAR-10. The VAE latent dimension of VAAL is 64. We begin the active learning cycle with 100 labeled instances, and in each cycle, we add another 100 labeled instances. The other settings are in accordance with CIFAR-10.

\begin{figure}[!htbp] \centering \includegraphics[width=3.2in]{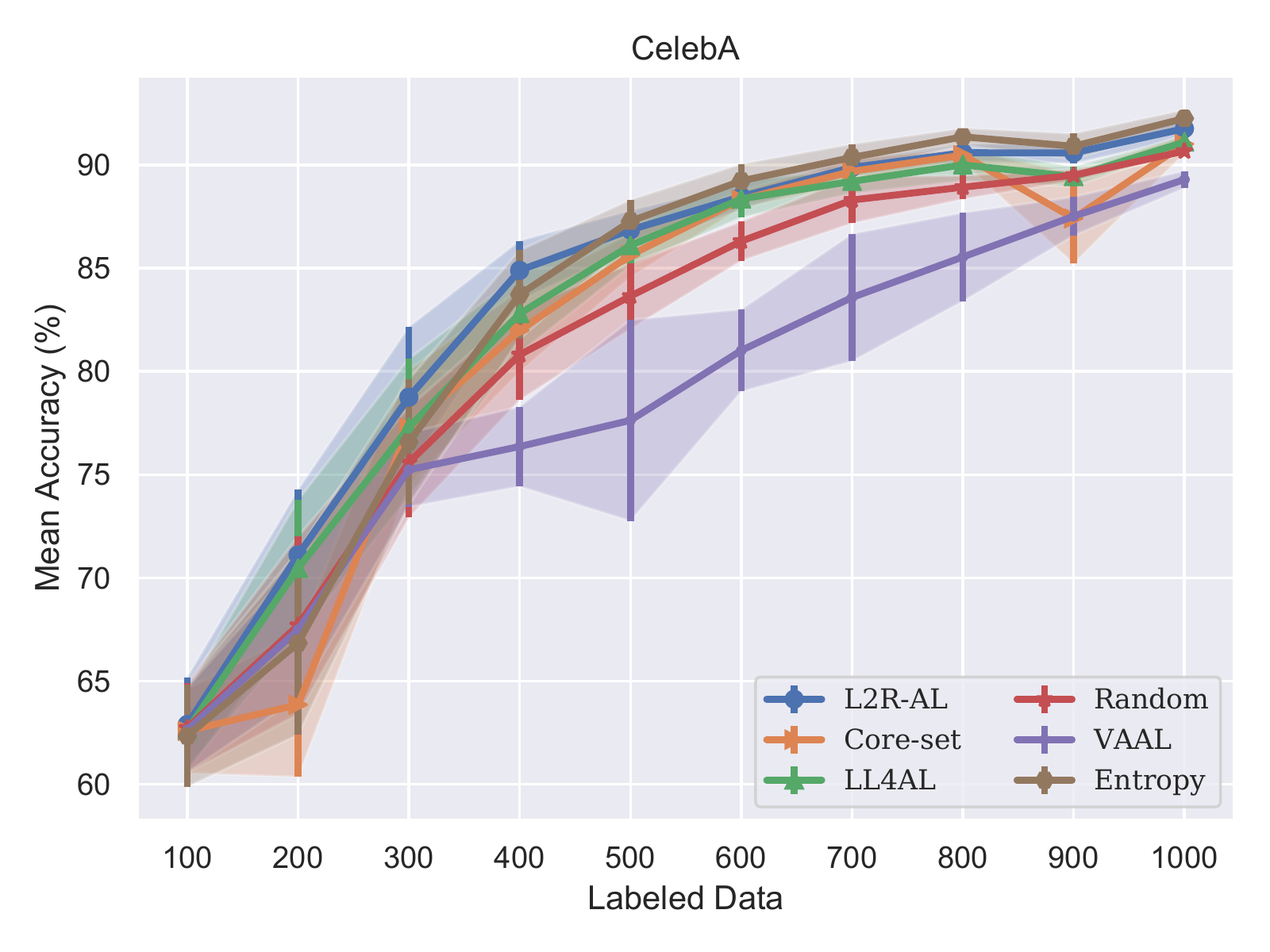}
\caption{Active learning results of CelebA image classification}
\label{celeba} 
\end{figure}

\noindent
\textbf{Results.} Fig.~\ref{celeba} shows the gender classification results on test set of 5 runs. Our algorithm is marked as the blue one and we can see it shows better results than the comparing algorithms before the fifth cycle and then entropy method performs better. In the last cycle, our algorithm achieves a 91.76\% result, compared with core-set's 91.02\%, LL4AL's 91.11\%, random approach's 90.67\%, VAAL's 89.29\% and entropy method's 92.25\%. The entropy method is  upper bound for this classification problem and our proposed algorithm outperforms the other algorithms.

The results of these two image classification tasks demonstrate that our algorithm outperforms the other state-of-the-art approaches and the random baseline. However, in the case of entropy, the results show it dominates all other active learning approaches (including ours). It's worth to mention this aspect, since entropy method is seldomly used as a baseline in the active learning literature. This could be explained by the fact that entropy is a very good measure for uncertainty of classification.

\subsection{Experiment on Human Pose Estimation}

\noindent
\textbf{Implementation details.} We implement active learning algorithms on human pose estimation problem and choose a simple baseline method \cite{xiao2018simple} for this regression problem.
In accordance with \cite{yoo2019learning}, we use the MPII dataset \cite{andriluka20142d} and follow the same splits: the training set consists of 22,246
poses from 14,679 images and the test set consists of 2,958 poses from 2,729 images. The subset size for using query strategy is 5000, which is same as \cite{yoo2019learning}. By default, the input size is cropped to $256 \times 256$ pixels. We use the default optimizer of their official code as shown in their paper \cite{xiao2018simple} to train the target model: an Adam optimizer with initial learning rate 1e-3. The target model and the loss prediction modules are trained for 140 epochs. We use the SGD optimizer to train the loss prediction modules with an initial learning rate of 5e-2, and after 110 epochs, it is decreased to 5e-3. We use the default target model which simply adds a few
deconvolutional layers over the last convolution stage in the ResNet-50. The features are extracted before the deconvolutional layers of the target model.

We use the standard evaluation metric for this problem: PCKh@0.5 \cite{andriluka20142d} which measures the percentage of predicted key-points falling within a threshold distance 0.5 to the ground truth, where the distance is normalized by a fraction of the head size. We begin with 200 labeled instances, and in each cycle, we add another 200 labeled instances, and stop training after 1000 labeled instances. The batch size is 32. The other settings are in accordance with the image classification tasks.

\noindent
\textbf{Comparing algorithms.} We compare our algorithm with core-set \cite{sener2018active}, LL4AL\cite{yoo2019learning} and two baselines: random method and entropy method \cite{li2013adaptive}. 

For entropy method, we use the same approach as \cite{yoo2019learning}: applying the softmax to each heatmap to get an entropy for each body part, then averaging all of the entropy values. 

\begin{figure}[!htbp] \centering \includegraphics[width=3.2in]{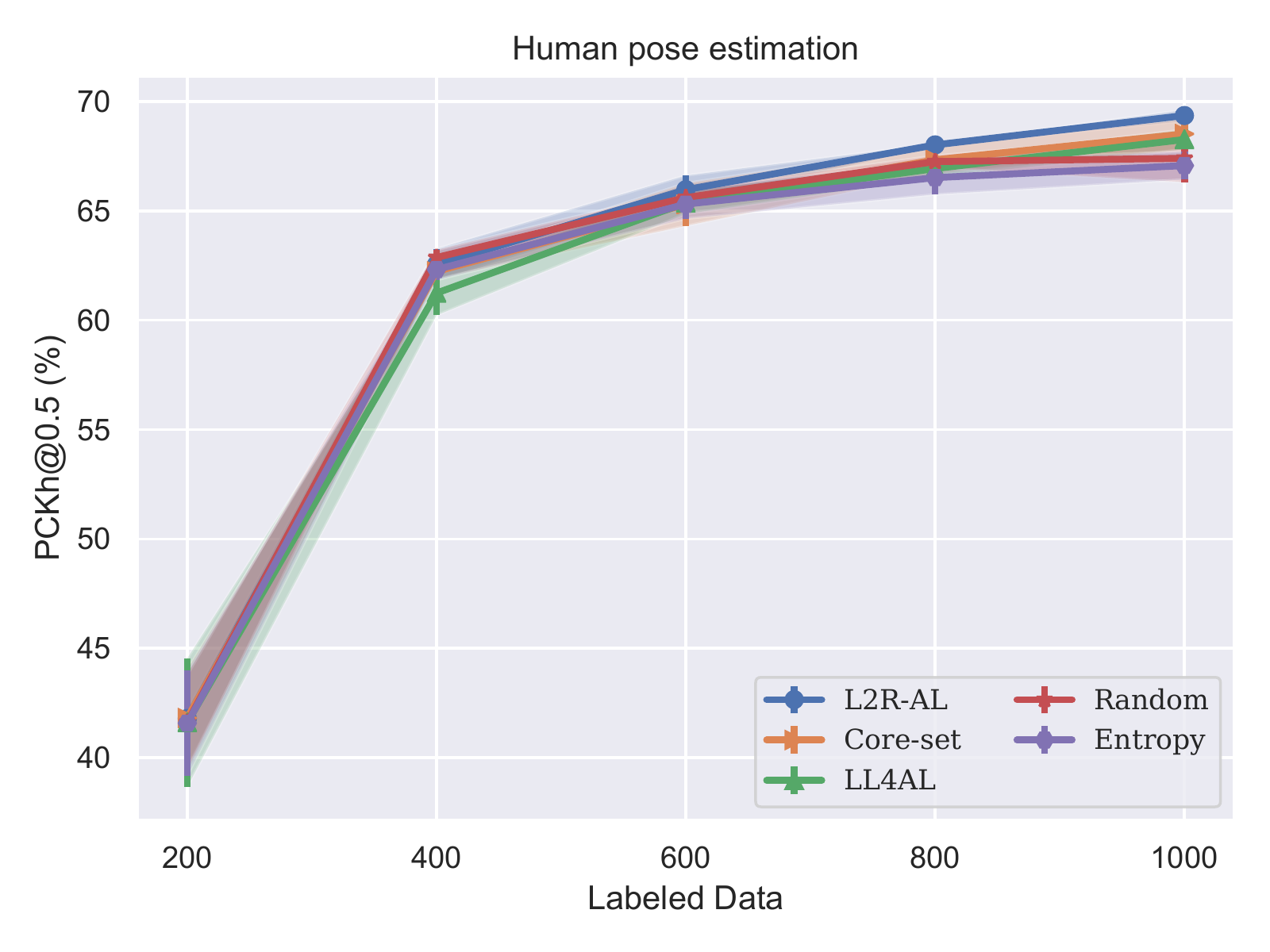}
\caption{Active learning results of human pose estimation}
\label{human} 
\end{figure}

\noindent
\textbf{Results.} Fig.~\ref{human} shows the average PCKh@0.5 comparisons of 5 runs. When the number of labeled instances is 600 or more, our algorithm achieves higher performance. In the last active learning cycle, our approach obtains 69.37\% result. By comparison, the core-set approach obtains 68.53\%, the LL4AL approach obtains 68.27\%, the random baseline obtains 67.41\%, and the entropy baseline achieves 67.08\%, respectively. Our algorithm achieves the highest result and the entropy method fails to query the informative instances for this problem.

\subsection{Experiment on Crowd Counting}

\noindent
\textbf{Implementation details.}
We implement active learning algorithms on crowd counting problem and choose an open source PyTorch code\footnote{\url{https://github.com/gjy3035/C-3-Framework}} for this regression problem. We use ShanghaiTech Part\_B \cite{zhang2016single} dataset, which has 400 and 316 images for training and testing, respectively. 

As the training set only has 400 images, the active learning algorithms query from the set directly without using subset. The target model and loss prediction module are trained for 100 epochs with the open source code's default optimizer: Adam optimizers with initial learning rate 1e-5. We choose the pretrained ResNet-50 based target model which is changed after the third layer with decoder and deonvolutional layers, as the authors \cite{gao2019c} have shown its good results. The features are extracted before the decoder layers of the target model and input to the loss prediction module.

We begin with 20 labeled instances, and in each cycle, we add another 20 labeled instances, and stop training after 100 labeled instances. The batch size is 4. The other settings are in accordance with the image classification problems.

\noindent
\textbf{Comparing algorithms.} 
Same as in the the experiment on human pose estimation and applying the softmax to the output density map for entropy method.

\begin{figure}[!htbp] \centering \includegraphics[width=3.2in]{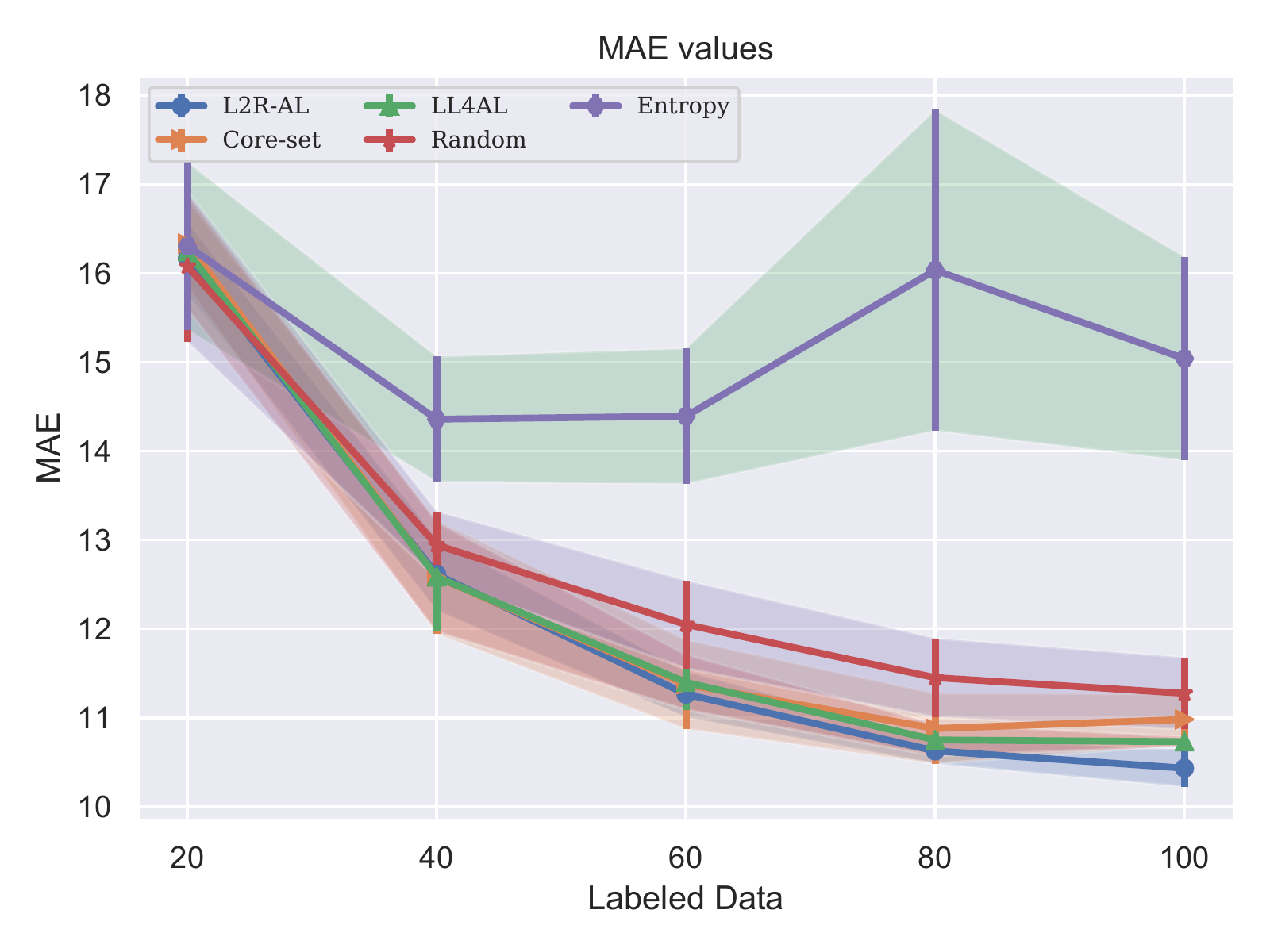}
\caption{Active learning results of crowd counting}
\label{crowdcount} 
\end{figure}

\noindent
\textbf{Results.}
Fig.~\ref{crowdcount} shows the comparisons in performance for the aforementioned algorithms, using MAE as evaluation metric. In this case, it measures the mean average error of predicted number and ground-truth number of persons. The result is an average over 5 runs. We can see that the proposed algorithm L2R-AL achieves lower errors than the comparing algorithms. In the last cycle, the MAE of our algorithm is 10.43, the LL4AL approach is 10.73, the core-set approach 10.98, the random approach is 11.28, and the entropy method is 15.04. 
We can see that our approach achieves the lowest error while the entropy method fails to query the informative instances too and is not effective for regression problems.

The above experiment results show that entropy method is still the outstanding approach for classification problems, but our algorithm also can match with it and outperforms the other algorithms. For regression problems, the entropy method fails and our algorithm achieves the best results.

\section{Conclusion}
\label{sec5}
In this paper, we propose a learning  loss based method for active learning which is task-agnostic: it attaches a module to learn to predict the target   loss of unlabeled data, and select data  with the highest loss for labeling. Furthermore, we demonstrate that learning loss based active learning algorithm actually is a learning to rank problem. We use a simple and effective listwise approach to train the loss prediction module by optimizing the Spearman's rank correlation metric. We validate the proposed approach on two tasks: image classification (CIFAR-10, CelebA) and regression (MPII, ShanghaiTech Part\_B). The experimental results show that our algorithm outperforms recent state-of-the-art active learning algorithms.


\section*{Acknowledgment}
The authors want to thank CERCA Program of Generalitat de Catalunya and Spanish project PID2019-104174GB-I00 (MINECO). 
Minghan Li acknowledges the Chinese Scholarship Council (CSC) grant No.201906960018.



\normalem
\bibliographystyle{IEEEtran}
\bibliography{IEEEexample}

%



\end{document}